\documentclass{article} 
\usepackage{iclr2027_conference,times}

\usepackage{amsmath,amsfonts,bm}

\def \reidrate{\text{ReID@1}}

\def\eqref#1{equation~\ref{#1}}

\def\1{\bm{1}}

\def\ra{{\textnormal{a}}}

\def\rx{{\textnormal{x}}}

\def\rva{{\mathbf{a}}}

\def\erva{{\textnormal{a}}}

\def\ervx{{\textnormal{x}}}

\def\rmA{{\mathbf{A}}}

\def\vmu{{\bm{\mu}}}
\def\vtheta{{\bm{\theta}}}
\def\va{{\bm{a}}}

\def\ve{{\bm{e}}}

\def\vx{{\bm{x}}}

\def\eva{{a}}

\def\mA{{\bm{A}}}

\def\mH{{\bm{H}}}
\def\mI{{\bm{I}}}
\def\mJ{{\bm{J}}}

\def\mX{{\bm{X}}}

\def\mSigma{{\bm{\Sigma}}}

\DeclareMathAlphabet{\mathsfit}{\encodingdefault}{\sfdefault}{m}{sl}
\SetMathAlphabet{\mathsfit}{bold}{\encodingdefault}{\sfdefault}{bx}{n}
\newcommand{\tens}[1]{\bm{\mathsfit{#1}}}
\def\tA{{\tens{A}}}

\def\tX{{\tens{X}}}

\def\gG{{\mathcal{G}}}

\def\sA{{\mathbb{A}}}
\def\sB{{\mathbb{B}}}

\def\sS{{\mathbb{S}}}

\def\emA{{A}}

\newcommand{\etens}[1]{\mathsfit{#1}}

\def\etA{{\etens{A}}}

\newcommand{\E}{\mathbb{E}}

\newcommand{\R}{\mathbb{R}}

\newcommand{\KL}{D_{\mathrm{KL}}}
\newcommand{\Var}{\mathrm{Var}}

\newcommand{\Cov}{\mathrm{Cov}}

\newcommand{\normltwo}{L^2}
\newcommand{\normlp}{L^p}

\newcommand{\parents}{Pa} 

\usepackage{hyperref}
\usepackage{url}
\usepackage{ifthen}
\usepackage{amsmath,amssymb}
\usepackage{graphicx}
\usepackage{booktabs}
\usepackage{amssymb}
\usepackage{comment}
\usepackage{tikz}
\usepackage{subcaption}
\usepackage{wrapfig}

\usetikzlibrary{arrows.meta,calc,positioning,fit,backgrounds}

\title{Quantifying privacy risks in foundation model representations of brain MRI}

\author{%
  Imen Bajar\textsuperscript{1,2} \quad
  Héber H. Arcolezi\textsuperscript{1} \quad
  Antoine Boutet\textsuperscript{2} \quad
  Carole Frindel\textsuperscript{3,4} \quad
  Mohamed Maouche\textsuperscript{2} \\[0.6em]
  \textsuperscript{1}École de technologie supérieure (ÉTS), Montréal, QC, Canada \\
  \textsuperscript{2}Inria, INSA Lyon, CITI, UR3720, 69621 Villeurbanne, France \\
  \textsuperscript{3}CREATIS, CNRS UMR5220, INSERM U1294, Université Lyon, INSA-Lyon, Lyon, France \\
  \textsuperscript{4}Institut Universitaire de France (IUF), Paris, France \\[0.4em]
}

 \newbool{showcomments}
 \setbool{showcomments}{true}
 \ifthenelse{\boolean{showcomments}}
 { \newcommand{\mynote}[3]{
    \fbox{\bfseries\sffamily\scriptsize#1}{\small$\blacktriangleright$\textsf{\emph{\color{#3}{#2}}}$\blacktriangleleft$}
    }}
 { \newcommand{\mynote}[4]{}}

\newcommand{\imen}[1]{\mynote{Imen}{#1}{blue}}

\newcommand{\antoine}[1]{\mynote{Antoine}{#1}{brown}}
\newcommand{\done}[1]{%
  \ifthenelse{\boolean{showcomments}}%
    {{\color{blue}[DONE#1]}}%
    {}%
}

\iclrfinalcopy 
\begin{document}

\maketitle

\begin{abstract}
Pretrained representations are increasingly used to share and reuse information from medical images without exposing the original scans, but it is unclear how much patient-specific information these representations retain. We investigate this question for brain magnetic resonance imaging (MRI) by evaluating five frozen pretrained encoders spanning different architectures, pretraining objectives, and training domains on two longitudinal T1-weighted datasets, OASIS-2 and MIRIAD. Under an attacker model with access to released embeddings, the corresponding public encoder, and auxiliary MRI scans, we evaluate patient linkage and sensitive-attribute inference. Across the evaluated representations, we observe substantial subject-specific information: rank-1 patient linkage reaches up to 99\% on MIRIAD and exceeds 90\% on OASIS-2 for the highest-risk representation, despite the absence of an identification objective during encoder training. We further show that the released embeddings retain information about a demographic attribute, with recorded-sex prediction reaching balanced accuracy of 90.9\% on OASIS-2 and 95.5\% on MIRIAD. Overall, our results demonstrate that replacing raw brain MRI with learned embeddings does not, by itself, eliminate patient-linkage or attribute-inference risk, and is therefore insufficient on its own to establish anonymity.

\end{abstract}

\section{Introduction}
\label{sec:intro}

Magnetic resonance imaging (MRI) is an imaging modality that generates detailed images of the internal structures of the body, including the brain's anatomy. Given this level of anatomical detail, brain MRI is a routine tool in the diagnosis and monitoring of neurological and psychiatric conditions such as Alzheimer's disease, stroke, and brain tumors~\citep{alzheimer,stroke,braintumors}. Many AI models have been trained to detect and classify these conditions from brain MRI scans ~\citep{7950647,mohsen2018classification}. 

At the same time, studies have shown that brain morphology can serve as a biometric signature, a ``brain fingerprint'', which enables subject re-identification through comparison with reference data~\citep{Wachinger2014,takao2015brain,valizadeh2018}.
Therefore, brain MRI has a dual nature, at once a diagnostic tool and a carrier of identifying biometric information.

In recent years, foundation models, large neural networks trained on broad and diverse datasets, have emerged. Many of these models use self-supervised learning to learn transferable representations that can be adapted to a range of downstream tasks~\citep{bommasani2022opportunitiesrisksfoundationmodels}. This approach is especially useful for medical imaging, where labeled data can be scarce and costly to obtain. Instead of training a model from scratch, a pretrained model can be adapted to the specific task at hand with limited task-specific labeled data~\citep{liu2025modalityagnosticmultitaskfoundationmodel,tak2026generalizable}.
For brain MRI specifically, a growing body of work has explored foundation models for representation learning and downstream analysis~\citep{barbano2026anatomical,ghamizi2025brainimagingfoundationmodels,mazher2025generalisablefoundationmodelsbrain}.
Throughout this paper, we refer to the vector representation produced by a pretrained encoder as an \emph{embedding}.

This question is also relevant to how anonymization is assessed in practice. Recent guidelines from the European Data Protection Board (EDPB) identify three criteria for evaluating whether data can be considered anonymous: \emph{No Record Isolation, No Linkage, and No Inference}~\citep{EDPB2026}. Our study examines two corresponding technical risks for released brain MRI representations: linkage across records belonging to the same individual and inference of withheld patient attributes. 

As pretrained encoders are increasingly used to produce reusable representations of brain anatomy for downstream tasks, a central privacy question arises: \textbf{\emph{when a learned representation is released instead of the raw brain MRI, can it still enable patient linkage or reveal additional patient-specific information?}}
Across the evaluated models and datasets, we observe substantial variation in privacy risk, with rank-1 patient matching exceeding 90\% for some representations and remaining well above chance across several model--dataset combinations.
We address this question through four contributions:
\begin{itemize}
    \item Define explicit threat models for patient linkage, re-identification, and sensitive-attribute inference from released brain MRI representations, and systematically evaluate five frozen pretrained encoders across two longitudinal MRI datasets.
    
    \item Quantify identity-related information through a score-based linkability audit and an explicit rank-1 matching attack (exploiting the proximity of embeddings derived from the same patient), showing that several representations preserve substantial subject-specific information despite the absence of an identification objective during encoder training.
    
    \item Evaluate sensitive-attribute inference from released brain MRI representations, showing that the embeddings can reveal the subject's recorded sex in addition to supporting patient linkage.
    
    \item Demonstrate that releasing embeddings instead of raw brain MRI does not, by itself, eliminate linkage or attribute-inference risk and therefore is insufficient, on its own, to establish anonymity.
\end{itemize}



\section{Related Work} 


\textbf{Privacy risks in brain MRI}: 
Historically, safeguarding patient privacy in medical imaging, as required by privacy regulations like the General Data Protection Regulation (GDPR)~\cite{european_union_2016_gdpr}, Personal Information Protection and Electronic Documents Act (PIPEDA)~\cite{PIPEDA}, and California Consumer Privacy Act (CCPA)~\cite{california_2026_ccpa}, has involved removing direct identifiers from medical data before secondary use or disclosure. In medical imaging, this typically includes stripping metadata such as names, birth dates, and medical record numbers from image headers. However, medical images themselves still contain detailed clinical and anatomical information, and patient-specific traces may remain even after obvious identifiers are removed~\citep{RSNA}.

Brain MRIs, for instance, capture detailed facial anatomy, which makes 3D facial reconstruction and therefore identification via face-recognition deep learning achievable. 
\citep{Boutet2023} demonstrated that faces reconstructed in 3D from brain MRI scans could be matched against public photos using deep learning, effectively exposing a person's identity. To guard against this, techniques like defacing are commonly used \citep{PyDeface2019, PyFaceWipe2024}. Yet defacing does not fully eliminate the risk either: deep learning can still reconstruct a plausible face from a defaced scan \citep{Jeong2020, Abramian2019}.

The problem is not limited to the face. Research has shown that the shape of the brain itself can serve as an identifying signature, a ``brain fingerprint''. 
\citet{Wachinger2014} developed BrainPrint, a technique that uses brain shape to identify individuals with near-perfect accuracy. 
\citet{valizadeh2018} found that brain features remain distinctive and stable even across years. 
DeepBrainPrint~\citep{Puglisi2023} explicitly trained a contrastive representation for longitudinal brain MRI re-identification, showing that learned representations can also encode stable subject-specific information across scans. 
More recently, \citet{Bajar} reinforced this finding by testing four defacing methods on brain scans; even after defacing, subjects were still re-identified 66–73\% of the time using brain-feature-based similarity attacks.
Taken together, these findings show that brain MRI can reveal identity in more than one way through brain features even after defacing methods have been applied.

\textbf{Medical foundation models}: 
Brain MRI foundation models can be broadly grouped according to their pretraining objective. Task-agnostic models are pretrained without a specific downstream task in mind; they rely on self-supervised objectives to learn general-purpose representations that can later be fine-tuned for various tasks~\citep{barbano2026anatomical, 3DNeuroSimCLR, brainmriSSL,caron2021emerging}.

Task-specific foundation models, in contrast, are pretrained around a particular downstream objective, most often segmentation~\citep{wang2024sammed3d, cox2024brainsegfounder3dfoundationmodels}. Their encoders are therefore shaped by this task rather than by the goal of producing general-purpose representations. We evaluate representative models from these different paradigms; their architectures and pretraining objectives are described in Section~\ref{sub:models}.

\textbf{Privacy risks for medical foundation models}:
~\citet{nebbia} showed that representations extracted from frozen medical foundation models can support patient retrieval in fundus photography, OCT, and chest X-ray, and can also expose demographic attributes. However, it does not define a threat model. Also, subsequent work questioned whether some of these results reflect properties specific to foundation models or simpler acquisition and image-level similarity, highlighting the importance of carefully specifying the attacker and evaluation protocol~\citep{Engelmann2026}. More broadly, learned representations are known to expose sensitive information beyond their intended downstream objective~\citep{song2019overlearning}, and recent work has emphasized the importance of explicit attacker knowledge and disclosure assumptions when evaluating clinical foundation models~\citep{tonekaboni2026protecting}. 

\textbf{Summary:} Our work connects two previously separate observations: brain anatomy is individually distinctive, acting as a biometric trait, and reusable learned representations can leak information beyond their intended purpose. Focusing on longitudinal brain MRI, we evaluate whether frozen pretrained encoders enable both patient linkage and sensitive-attribute inference, explicitly separating linkability from re-identification threat models. By quantifying how much these representations expose patients to re-identification, we highlight the need for stronger safeguards and greater awareness as such models are released as open-source resources and made widely accessible.

\section{Threat Model and Experimental Setup} 
\label{sec:exp-setup}

This section describes the experimental setting used to evaluate privacy risks in pretrained brain MRI representations. We first introduce the evaluated encoders and longitudinal datasets, then formalize the attacker capabilities and privacy threats, and finally describe the embedding extraction, linkage, attribute-inference, and evaluation protocols.

\subsection{Models and datasets} \label{sub:models}
\textbf{Models.} 
We evaluate five pretrained encoders that differ in pretraining domain, input dimensionality, architecture, and objective (Table~\ref{tab:brain-mri-models}): AnatCL~\citep{barbano2026anatomical}, 3D-Neuro-SimCLR~\citep{3DNeuroSimCLR}, Brain-MRI-SSL~\citep{brainmriSSL}, DINO-ViT~\citep{caron2021emerging}, and SAM-Med3D~\citep{wang2024sammed3d}. The first three were pretrained specifically on brain MRI, whereas DINO-ViT was pretrained on natural images and SAM-Med3D on heterogeneous medical images. AnatCL, 3D-Neuro-SimCLR, and SAM-Med3D process full 3D volumes, while Brain-MRI-SSL and DINO-ViT operate on 2D images. Most models are trained with self-supervised objectives to learn general-purpose representations, whereas SAM-Med3D is trained specifically for segmentation in a supervised way. The models also use different loss objectives for their training. For instance, AnatCL is trained with weakly contrastive loss that combines two terms: an anatomical loss that brings closer subjects with similar anatomical descriptors, and a y-Aware loss that does the same based on age. 3D-Neuro-SimCLR and Brain-MRI-SSL both use contrastive learning. 3D-Neuro-SimCLR uses SimCLR’s NT-Xent loss, which pulls together two views of the same scan and pushes apart different scans in a batch. Brain-MRI-SSL uses MoCov3's symmetrized InfoNCE loss: each augmented view, encoded by a gradient-updated encoder, is contrasted against the other view produced by a momentum-updated encoder, with the other samples of the batch serving as negatives. In contrast, DINO-ViT relies on self-distillation. It minimizes the cross-entropy between a student network and a momentum teacher, without using negatives or labels. Finally, SAM-Med3D uses a supervised segmentation loss (Dice and cross-entropy). This diversity allows us to assess whether privacy risks persist across substantially different pretraining domains, architectures, and objectives.

\begin{table}[ht]
  \centering
  \small
  \setlength{\tabcolsep}{4pt}
  \resizebox{\textwidth}{!}{
  \begin{tabular}{lllllr}
    \toprule
    Model & Pretraining objective & Input & Backbone & Pretraining domain & Dim. \\
    \midrule
    AnatCL~\citep{barbano2026anatomical}      & Weakly sup.\ contrastive & 3D & ResNet18 & Brain MRI           & 512  \\
    3D-Neuro-SimCLR~\citep{3DNeuroSimCLR}     & Contrastive (SimCLR)     & 3D & ResNet18 & Brain MRI           & 512  \\
    Brain-MRI-SSL~\citep{brainmriSSL}         & Contrastive (MoCov3)       & 2D & ResNet50 & Brain MRI           & 2048 \\
    DINO-ViT~\citep{caron2021emerging}        & Self-distillation        & 2D & ViT-B/8  & Natural images      & 768  \\
    SAM-Med3D~\citep{wang2024sammed3d}        & Sup. segmentation             & 3D & ViT      & Multi-organ medical & 384  \\
    \bottomrule
  \end{tabular}
   }
    \caption{Overview of the pretrained encoders evaluated in this work.}
  \label{tab:brain-mri-models}
\end{table}


\textbf{Datasets.} We use two T1-weighted brain MRI datasets that were not included in the training data of any of our tested models: OASIS-2~\citep{Marcus2010} and MIRIAD~\citep{malone2013miriad}, summarized in Table~\ref{tab:oasis-miriad}. Both are longitudinal datasets including Alzheimer's patients and healthy controls, but they differ substantially in scale and acquisition design. OASIS-2 includes more subjects (150) but fewer scans per subject (2--5 sessions), whereas MIRIAD includes fewer subjects (69) but with more repeated scans per subject (up to 12 scans over 9 visits, 708 scans in total). OASIS-2 scans are provided in a preprocessed form (motion-corrected, atlas-registered, skull-stripped, and intensity-corrected). Importantly, these are longitudinal acquisitions rather than near-duplicate acquisitions of the same scan: MIRIAD follow-ups are separated by approximately two weeks to two years, while OASIS-2 follow-ups extend over multiple years. The attack therefore evaluates whether subject-specific information persists across distinct acquisitions over time.

\begin{table}[ht]
\centering
\vspace{2mm}
\begin{tabular}{|l|c|c|}
\hline
\textbf{Characteristics} & \textbf{OASIS-2} & \textbf{MIRIAD} \\
\hline
Subjects & 150 & 69  \\
\hline
Sex & 58.7\% F, 41.3\% M & 55.1\% F, 44.9\% M  \\
\hline
Classes & 72 NC, 64 AD, 14 Converted & 46 AD, 23 CN \\
\hline
Preprocessing & Yes & No \\
\hline
Scans per subject & \shortstack{at least 2 scans \\ variable follow-ups (1--6+ years)} & \shortstack{4--12 scans (up to 9 visits) \\ intervals of 2 weeks--2 years} \\
\hline
\end{tabular}
\vspace{2pt}
\begin{minipage}{0.95\linewidth}
\footnotesize
\vspace{2mm}
\textbf{Abbreviations:} AD = Alzheimer's Disease, MCI = Mild Cognitive Impairment, CN = Cognitively Normal, NC = Nondemented Controls. F = Female, M=Male.
\end{minipage}
\caption{Characteristics of the OASIS-2 and MIRIAD datasets used in this study.}
\vspace{-2mm}
\label{tab:oasis-miriad}
\end{table}

\subsection{Threat Model}
\label{sub:threat_model}

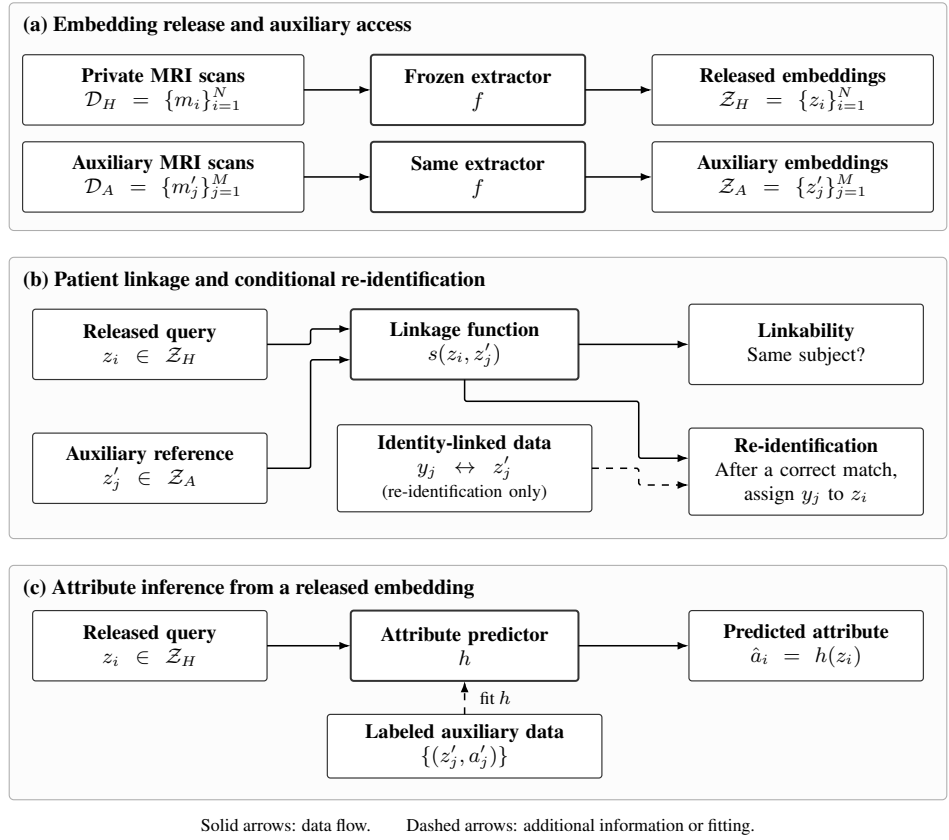
\begin{figure}[t]
\centering
\resizebox{0.9\linewidth}{!}{
\begin{tikzpicture}[
x=1mm, y=1mm,
font=\fontsize{8.6}{10.2}\selectfont,
>=Latex,
box/.style={
draw=black!80, line width=.5pt, rounded corners=1.3pt,
fill=white, align=center, text width=33mm,
minimum height=10.5mm, inner xsep=1mm, inner ysep=.7mm,
outer sep=0pt
},
model/.style={box, line width=.85pt},
flow/.style={-{Latex[length=1.7mm,width=1.2mm]}, line width=.65pt, rounded corners=1pt},
extra/.style={flow, dashed},
heading/.style={anchor=north west, inner sep=0pt, font=\fontsize{9}{10.5}\selectfont\bfseries},
note/.style={font=\fontsize{7.6}{9}\selectfont, align=center},
panel/.style={draw=black!32, line width=.45pt, rounded corners=2pt, fill=black!1}
]
\begin{scope}[on background layer]
\draw[panel] (0,0) rectangle (140,-34);
\end{scope}
\node[heading] at (2,-2) {(a) Embedding release and auxiliary access};
\node[box,text width=40mm] (hospital) at (23,-13) {\textbf{Private MRI scans}\\$\mathcal D_H=\{m_i\}_{i=1}^{N}$};
\node[model,text width=30mm] (fholder) at (70,-13) {\textbf{Frozen extractor}\\$f$};
\node[box,text width=40mm] (released) at (117,-13) {\textbf{Released embeddings}\\$\mathcal Z_H=\{z_i\}_{i=1}^{N}$};
\draw[flow] (hospital.east) -- (fholder.west);
\draw[flow] (fholder.east) -- (released.west);
\node[box,text width=40mm] (auxiliary) at (23,-26) {\textbf{Auxiliary MRI scans}\\$\mathcal D_A=\{m'_j\}_{j=1}^{M}$};
\node[model,text width=30mm] (faux) at (70,-26) {\textbf{Same extractor}\\$f$};
\node[box,text width=40mm] (auxemb) at (117,-26) {\textbf{Auxiliary embeddings}\\$\mathcal Z_A=\{z'_j\}_{j=1}^{M}$};
\draw[flow] (auxiliary.east) -- (faux.west);
\draw[flow] (faux.east) -- (auxemb.west);
\begin{scope}[on background layer]
\draw[panel] (0,-38) rectangle (140,-80);
\end{scope}
\node[heading] at (2,-40) {(b) Patient linkage and conditional re-identification};
\node[box] (query) at (21,-51) {\textbf{Released query}\\$z_i\in\mathcal Z_H$};
\node[box] (reference) at (21,-69.5) {\textbf{Auxiliary reference}\\$z'_j\in\mathcal Z_A$};
\node[model,text width=32mm] (score) at (68,-51) {\textbf{Linkage function}\\$s(z_i,z'_j)$};
\node[box,minimum height=12mm] (linkability) at (119,-51) {\textbf{Linkability}\\Same subject?};
\node[box,minimum height=13mm] (reid) at (119,-70) {\textbf{Re-identification}\\After a correct match,\\assign $y_j$ to $z_i$};
\node[box,text width=36mm,minimum height=13mm] (identity) at (68,-69.5) {\textbf{Identity-linked data}\\$y_j\leftrightarrow z'_j$\\{\fontsize{7.6}{9}\selectfont (re-identification only)}};
\coordinate (scoreTop) at ([yshift=2.3mm]score.west);
\coordinate (scoreBottom) at ([yshift=-2.3mm]score.west);
\draw[flow] (query.east) -- (45,-51) |- (scoreTop);
\draw[flow] (reference.east) -- (45,-69.5) |- (scoreBottom);
\draw[flow] (score.east) -- (linkability.west);
\coordinate (reidTop) at ([yshift=2mm]reid.west);
\coordinate (reidBottom) at ([yshift=-2mm]reid.west);
\draw[flow] (score.south) -- (68,-59.5) -- (94,-59.5) |- (reidTop);
\draw[extra] (identity.east) -- (94,-69.5) |- (reidBottom);
\begin{scope}[on background layer]
\draw[panel] (0,-84) rectangle (140,-119);
\end{scope}
\node[heading] at (2,-86) {(c) Attribute inference from a released embedding};
\node[box] (attributeQuery) at (21,-96) {\textbf{Released query}\\$z_i\in\mathcal Z_H$};
\node[model,text width=32mm] (attributeModel) at (68,-96) {\textbf{Attribute predictor}\\$h$};
\node[box] (attributeOutput) at (119,-96) {\textbf{Predicted attribute}\\$\hat a_i=h(z_i)$};
\node[box,text width=39mm,minimum height=9.5mm] (attributeTrain) at (68,-111) {\textbf{Labeled auxiliary data}\\$\{(z'_j,a'_j)\}$};
\draw[flow] (attributeQuery.east) -- (attributeModel.west);
\draw[flow] (attributeModel.east) -- (attributeOutput.west);
\draw[extra] (attributeTrain.north) -- (attributeModel.south) node[midway,right=1mm,note] {fit $h$};
\node[note,anchor=center] at (70,-123) {Solid arrows: data flow.\qquad Dashed arrows: additional information or fitting.};
\end{tikzpicture}
\vspace{-1mm}
}
\caption{Threat models for released brain MRI representations. (a) The data holder and attacker apply the same frozen feature extractor $f$ to private and auxiliary scans, respectively; either collection may contain multiple acquisitions per subject. The hospital's original scans remain unavailable to the attacker. (b) A linkage function assesses same-subject correspondence; re-identification additionally requires identity-linked reference information. (c) A predictor fitted to attribute-labeled auxiliary embeddings estimates a withheld attribute from a released embedding, without requiring successful linkage.}
\label{fig:threat_model}
\vspace{-6mm}
\end{figure}

Figure~\ref{fig:threat_model} summarizes three threats associated with released brain MRI embeddings: patient linkability, re-identification, and attribute inference.

\textbf{Released data and attacker access.} For each evaluated model, let $f$ denote its scan-to-embedding pipeline, comprising a publicly available frozen encoder and the preprocessing and aggregation operations described in Section~\ref{sub:eval_protocol}. Each scan $m$ is mapped to an embedding $f(m)\in\mathbb{R}^{d}$, where $d$ is the representation dimension for that model. A data holder has a private MRI collection $\mathcal{D}_H=\{m_i\}_{i=1}^{N}$ and releases only its embeddings $\mathcal{Z}_H=\{z_i\}_{i=1}^{N}$, with $z_i=f(m_i)$. The attacker obtains $\mathcal{Z}_H$ through an intended release or a data leak, but does not receive the underlying scans or their associated identities and attribute labels. The attacker knows $f$ and has auxiliary scans $\mathcal{D}_A=\{m'_j\}_{j=1}^{M}$, from which they compute $\mathcal{Z}_A=\{z'_j\}_{j=1}^{M}$, with $z'_j=f(m'_j)$. Throughout the paper, we use the terms \emph{embedding} and \emph{representation} interchangeably to refer to $z_i=f(m_i)$.

Both collections may contain multiple acquisitions per individual: $N$ and $M$ count scans, not distinct subjects. For linkage and re-identification, the collections may overlap in subjects through different acquisitions, without giving the attacker access to the hospital's original scans. Such overlap could arise, for example, when a patient undergoes MRI examinations at multiple clinics and an attacker with access to scans from one clinic attempts to link them to embeddings released by another. Let $u_i$ and $u'_j$ denote the ground-truth subject labels of $m_i$ and $m'_j$ in a common label space. These labels establish matching correctness during evaluation; they are not supplied for released queries. Encoder weights remain fixed, although attack models may be fitted using labeled auxiliary data.

\textbf{Linkability.} Given a released embedding $z_i$ and an auxiliary embedding $z'_j$, the attacker computes a similarity score $s(z_i,z'_j)$ to assess whether the corresponding scans originate from the same individual. The goal is to establish a correspondence between records belonging to the same subject, without requiring knowledge of that subject's real-world identity.

\textbf{Re-identification.} Re-identification builds on the same linkage procedure but assumes that the auxiliary reference records are additionally associated with identity-linked information $y_j$. The attacker ranks the auxiliary references according to their similarity to a released query $z_i$ and assigns the identity associated with the highest-scoring reference. Thus, the underlying matching operation is the same in both settings: linkability establishes that two records correspond to the same subject, whereas re-identification uses identity-linked auxiliary information to associate the released record with an identified individual.

\textbf{Attribute inference.} The attacker fits a predictor $h$ using attribute-labeled auxiliary embeddings $\{(z'_j,a'_j)\}$ and estimates the withheld attribute $a_i$ of a released scan as $\hat a_i=h(z_i)$. We evaluate inference of the subject's recorded sex. This attack requires auxiliary attribute labels, but neither identity-linked references nor a successful match to another acquisition from the query subject.
These threats concern information retained in released representations and do not require the corresponding scans to have been used to pretrain the encoder.

\subsection{Evaluation Protocol}
\label{sub:eval_protocol}

\textbf{Embedding extraction.} We apply model-specific preprocessing as described in its paper and extract one embedding per MRI scan. For Brain-MRI-SSL, we select 20 axial slices at fixed relative positions within the 25\%--75\% depth range of each volume. For DINO-ViT, we select 10 slices in each of the axial, coronal, and sagittal orientations, yielding 30 slices per volume within the same relative range along each axis. In both pipelines, slice embeddings are averaged within each volume. Thus, $f(m_i)$ is a scan-level representation, not an aggregation of different acquisitions from the same subject.

\textbf{Similarity scoring.} Let $\mathcal{Q}\subseteq\{1,\ldots,N\}$ denote the query indices and $\mathcal{G}_i\subseteq\{1,\ldots,M\}$ the eligible reference indices for query $i$. For each query embedding $z_i$, we compute a Canberra distance-based similarity score $s_{ij}=s(z_i,z'_j) = 1/\big(1+\sum_{k}\frac{|z_{ik}-z'_{jk}|}{|z_{ik}|+|z'_{jk}|}\big)$ against each eligible reference embedding $z'_j$, with larger scores indicating greater similarity. These scores are used to compute rank-1 linkage accuracy and estimate the same-subject and different-subject score distributions for the linkability measure. An ablation comparing Canberra and cosine similarities shows that Canberra gives the highest linkability in the evaluated setting (Appendix~\ref{app:ablation_distance_metric}); we therefore use it in the main analysis.
To conduct the experiment, each scan is treated in turn as the released query. Its reference gallery contains every other scan in the dataset; the query scan itself is excluded. Scans from the same subject constitute genuine references, while scans from all other subjects constitute impostor references. Since every subject has multiple longitudinal acquisitions, each query has at least one genuine reference.

\textbf{Attribute inference.} For each evaluated encoder and dataset, we train a random forest predictor $h$ to infer the subject's recorded sex from the extracted embeddings. We use subject-disjoint stratified cross-validation within each dataset: all scans from a given subject are assigned exclusively to either the training or held-out test, ensuring that no subject appears in both partitions. In each split, the training partition serves as the attribute-labeled auxiliary collection $\{(z'_j,a'_j)\}$, while the held-out partition serves as the collection of released query embeddings. The predictor is fitted to the auxiliary embedding--label pairs and applied to each held-out embedding $z_i$ to obtain $\hat a_i=h(z_i)$. The held-out attribute labels $a_i$ are used only to evaluate the predictions, and subject identifiers are not predictor inputs. No query--reference matching or Canberra scoring is used in this attack: attribute inference operates directly on the released embeddings through the fitted predictor (see Figure~\ref{fig:threat_model}). 

\textbf{Metrics.} We report $D_{\leftrightarrow}^{\mathrm{sys}}$ as a score-based audit of linkage risk, rank-1 accuracy as the success rate of the explicit matching attack, and held-out classification performance for attribute inference.

\emph{Linkability score.} We compute the system-level measure $D_{\leftrightarrow}^{\mathrm{sys}}$~\citep{Gomez2018,maouche2020comparative}. Let $S$ denote the similarity score, $H$ the same-subject hypothesis, and $\overline H$ the different-subject hypothesis, with prior $\pi=P(H)$. The posterior $P(H\mid S=t)$ is obtained from $p(t\mid H)$ and $p(t\mid\overline H)$ by Bayes' rule. The linkability measure is
\begin{equation}
D_{\leftrightarrow}^{\mathrm{sys}}=\int_{\mathbb{R}}p(t\mid H)\left[P(H\mid S=t)-P(\overline H\mid S=t)\right]_+\,\mathrm{d}t,
\label{eq:d-sys}
\end{equation}
where $[v]_+=\max(v,0)$. It lies in $[0,1]$, with larger values indicating stronger same-subject separability under the specified scoring procedure. A data holder with repeated acquisitions and known subject correspondence can compute this measure before releasing embeddings; we therefore use it as an audit metric rather than as the outcome of the explicit matching attack. Its value remains dependent on the scoring function, prior, and density estimation. We use all available same-subject and different-subject pairs to estimate the corresponding similarity-score distributions. Using the full set of available pairs provides the most complete empirical estimate of these distributions for each dataset, which are then used to compute the system-level linkability measure $D_{\leftrightarrow}^{\mathrm{sys}}$.

\emph{Rank-1 linkage accuracy (re-identification rate).} We evaluate the explicit matching attack by ranking the eligible auxiliary references for each query and selecting the highest-scoring reference, $\hat j(i)\in\operatorname*{arg\,max}_{j\in\mathcal{G}_i}s_{ij}$. We report
\begin{equation}
\reidrate=\frac{1}{|\mathcal{Q}|}\sum_{i\in\mathcal{Q}}\mathbf{1}\!\left\{u_i=u'_{\hat j(i)}\right\},
\label{eq:reid-rate}
\end{equation}
This scan-averaged metric measures successful subject linkage; its interpretation as re-identification additionally requires the identity-linked reference setting of Section~\ref{sub:threat_model}. In the experiments, each scan is treated once as the released query and compared with every other scan in the dataset. The reported \reidrate is therefore the proportion of individual scans for which the highest-scoring reference belongs to the same subject, averaged over all scans in the dataset.

\emph{Attribute-inference performance.} We report balanced accuracy and precision on held-out embeddings. Balanced accuracy gives equal weight to both recorded sex categories, with 
a performance of $0.5$ for random guess for this binary task. Higher performance indicates that the released representations retain information predictive of the withheld attribute.

\section{Privacy risk evaluation of pretrained brain representations}
We now evaluate the privacy risks defined in Section~\ref{sub:threat_model}. We first assess patient linkage and re-identification across the five pretrained encoders, then evaluate sensitive-attribute inference from the released embeddings. Finally, we discuss how differences in pretraining objectives may relate to the observed privacy risks.

\subsection{Patient Linkage and Re-identification}

\begin{table}[ht]
\centering
\setlength{\tabcolsep}{5pt}
\begin{tabular}{lcc|cc}
    \toprule
    & \multicolumn{2}{c|}{\textbf{OASIS-2}} & \multicolumn{2}{c}{\textbf{MIRIAD}} \\
    \cmidrule(lr){2-3} \cmidrule(lr){4-5}
    Model & Linkability & \reidrate & Linkability & \reidrate \\
    \midrule
    AnatCL             & 0.9312 & 0.9205 & 0.9714 & 0.9971 \\
    3D-Neuro-SimCLR    & 0.6571 & 0.6647 & 0.7938 & 0.9787 \\
    Brain-MRI-SSL      & 0.5647 & 0.5764 & 0.8460 & 0.9816 \\
    DINO-ViT           & 0.7796 & 0.8235 & 0.8761 & 0.9870 \\
    SAM-Med3D          & 0.2627 & 0.1394 & 0.5139 & 0.8958 \\
    \midrule
    FreeSurfer features (Baseline) & 0.8134 & 0.7500 & 0.9797 & 0.9993 \\
    \textit{Subject-level Random guess} & -- & \textit{0.0068} & -- & \textit{0.0145} \\
    \bottomrule
\end{tabular}
\caption{Re-identification risk evaluation of the pretrained encoders on OASIS-2 and MIRIAD}
\vspace{2mm}
\label{tab:reid-results}
\end{table}
Table~\ref{tab:reid-results} reports the results and some examples of the linkability plots are displayed in Appendix~\ref{app:linkability}. For MIRIAD, the risk of re-identifying patients is high across all models, with both a high re-identification rate and high linkability. SAM-Med3D shows the lowest risk, with a re-identification rate above 89\% and a linkability of 0.5139, which remains substantial. The other four models achieve a re-identification rate of at least 98\% and a linkability of at least 0.79, with AnatCL and Brain-MRI-SSL showing the highest values and therefore being the least private. On OASIS-2, results vary more across models: AnatCL and DINO-ViT are the riskiest, with linkabilities of 0.9312 and 0.7796 and re-identification rates of 92.05\% and 82.35\%, respectively, while SAM-Med3D is again the safest, with a linkability of 0.2627 and a re-identification rate of only 13.94\%.

Although none of these models was trained for re-identification, their embeddings enable it, and most of them are therefore not privacy-safe. The lower risk of SAM-Med3D may be related to its training objective: it is the only model trained in a supervised manner for organ segmentation, so its embeddings are optimized for a different purpose. All the other models are self-supervised and not trained with a specific downstream task in mind. DINO-ViT and SAM-Med3D are the only two models not specific to brain MRI, and DINO-ViT is not even trained on medical images. Surprisingly, it still poses a considerable re-identification risk, higher on OASIS-2 than brain-specific models such as Brain-MRI-SSL and 3D-Neuro-SimCLR.

Preprocessing may also contribute to these differences. Among the 3D models, AnatCL uses the most detailed pipeline: after nonlinear registration, it operates on gray matter maps, which describe the amount of gray matter at each location of the brain and therefore capture individual structural variations. 3D-Neuro-SimCLR uses skull-stripped and linearly registered volumes, and SAM-Med3D applies no brain-specific preprocessing. On OASIS-2, the risk follows this order, with a re-identification rate of 92.05\% for AnatCL, 66.47\% for 3D-Neuro-SimCLR, and 13.94\% for SAM-Med3D. On MIRIAD, AnatCL and 3D-Neuro-SimCLR both reach near-complete re-identification, while SAM-Med3D remains the least risky.

AnatCL is among the riskiest models on both datasets (re-identification rate above 0.9). Performing re-identification using only the three descriptors, guiding its weakly contrastive training, yields results close to those obtained with the full AnatCL embeddings, suggesting that these anatomical features drive much of its risk. Overall, these results show that pretrained brain MRI encoders are not privacy-safe: their embeddings can be used to re-identify patients, even though none of these models was trained for this purpose. For a fair comparison with OASIS-2, we repeated the MIRIAD analysis with 2--5 scans per subject (Appendix~\ref{app:nb-scans}).

This linkability comparison can serve as a practical audit, allowing a hospital to decide whether a model requires additional protection before using it or sharing its embeddings.  Risk varies across models: critical for AnatCL (above 0.93 on both datasets), consistently high for DINO-ViT (0.78--0.88), dataset-dependent for 3D-Neuro-SimCLR and Brain-MRI-SSL (0.79 and 0.85 on MIRIAD vs.\ 0.66 and 0.56 on OASIS-2), and lowest for SAM-Med3D, which still reaches 0.51 on MIRIAD and cannot be considered safe. 

We also evaluate an ad-hoc baseline representation following~\cite{Bajar}, which consists of 15 features derived from FreeSurfer rather than a learned embedding. On MIRIAD, this representation reaches a linkability of 0.9797 and a re-identification rate of 99.93\%. On OASIS-2, it remains highly identifying, with a linkability of 0.8134 and a re-identification rate of 75.00\%. These 15 features are manually extracted anatomical measurements known to be highly distinctive. Yet deep learning models can extract more discriminative features unintentionally and reach similar and sometimes higher re-identification risk.

\subsection{Sensitive Attribute Inference}
\label{sec:inference}

\begin{wraptable}{r}{0.625\textwidth}
\vspace{-0.8em}
\centering
\setlength{\tabcolsep}{2.pt}
\centering
\begin{tabular}{lcccc}
    \toprule
     & \multicolumn{2}{c}{OASIS-2} & \multicolumn{2}{c}{MIRIAD} \\
    \cmidrule(lr){2-3} \cmidrule(lr){4-5}
    Model            & Bal. Acc. & Precision & Bal. Acc. & Precision \\
    \midrule
    AnatCL           & 0.6772 & 0.6965 & 0.6946 & 0.6983 \\
    3D-Neuro-SimCLR  & 0.7011 & 0.7075 & 0.7789 & 0.7806 \\
    Brain-MRI-SSL    & 0.9093 & 0.9153 & 0.9546 & 0.9568 \\
    DINO-ViT         & 0.7894 & 0.8100 & 0.8716 & 0.8704 \\
    SAM-Med3D        & 0.8632 & 0.8735 & 0.8850 & 0.8887 \\
    \bottomrule
\end{tabular}
\caption{Sex classification performance on OASIS-2 and MIRIAD.}
\label{tab:sex-classification}
\vspace{-0.8em}
\end{wraptable}

Beyond re-identification, we investigate whether these model embeddings also leak demographic attributes, specifically sex. We train a Random Forest classifier on top of the extracted embeddings to predict the subject's sex. The classifier is evaluated using subject-disjoint stratified cross-validation, and we report balanced accuracy and precision as our metrics (Table~\ref{tab:sex-classification}). High classification performance indicates that sex-discriminative information is embedded in the representation, even though the foundation model was never explicitly trained for this task. This supports the claim that these embeddings encode more identity-linked information than intended, raising a further type of privacy risk beyond re-identification alone. Importantly, linkage and attribute inference do not induce the same ordering across representations. This indicates that privacy risk is multidimensional: reducing subject linkability does not necessarily reduce other sensitive information retained by the representation. Appendix~\ref{app:sex_cross} reports the results when the classifier is trained on one dataset and tested on the other.


\subsection{Implications of pretraining objectives for privacy}
\label{sec:discussion_objectives}

The evaluated encoders differ substantially in how their representations are learned, yet none is explicitly optimized to match longitudinal acquisitions from the same patient. This contrasts with biometric representation learning, where identity consistency is part of the objective. For example, \citet{fallahi2026advancing} train EEG encoders using supervised metric learning that brings recordings from the same subject across sessions closer while separating different subjects. In our setting, patient identity is never provided as supervision.

The contrastive models also differ in their training mechanism. Both 3D-Neuro-SimCLR and Brain-MRI-SSL treat augmented views of the same scan as positives and other scans as negatives, but 3D-Neuro-SimCLR optimizes an NT-Xent objective with a single shared encoder~\citep{3DNeuroSimCLR}, whereas Brain-MRI-SSL follows MoCov3, in which keys are produced by a momentum encoder~\citep{brainmriSSL}. These objectives are instance-based rather than identity-aware and do not explicitly encourage different acquisitions of the same subject to be close. If such acquisitions occur among the negatives, the objective may even discourage their similarity, depending on the pretraining sampling strategy. AnatCL instead guides contrastive learning using anatomical descriptors and age~\citep{barbano2026anatomical}; DINO-ViT uses self-distillation across augmented views~\citep{caron2021emerging}; and SAM-Med3D is optimized for supervised segmentation~\citep{wang2024sammed3d}.

Despite these different learning signals, all evaluated representations retain subject-specific information supporting longitudinal linkage, although the magnitude varies across models and datasets. This suggests that such leakage is not specific to a particular contrastive formulation or the use of explicit negatives, but can arise as a by-product of preserving stable anatomical, structural, or instance-specific characteristics. AnatCL provides a direct example: although not designed for identity matching, its anatomical descriptors alone already support strong re-identification. However, because the encoders also differ in architecture, pretraining data, and preprocessing, differences in linkability cannot be attributed to the training objective alone.

These findings have implications for both privacy-oriented representation learning and AI governance. An identity-agnostic objective should not be assumed to produce an identity-invariant representation, motivating privacy-aware training or post-hoc sanitization of released embeddings. For high-risk AI systems, Article~9 of the EU AI Act requires an iterative process for identifying and evaluating reasonably foreseeable risks to health, safety, and fundamental rights and adopting appropriate risk-management measures~\citep{EUAIAct2024}. Privacy audits such as the linkage and attribute-inference evaluations considered here can provide technical evidence of residual risks before deployment or release, complementing the data-protection assessment of linkage and inference risks discussed earlier~\citep{EDPB2026}, without by themselves establishing regulatory compliance or anonymity.

\section{Conclusion}
We audited whether released representations from frozen pretrained encoders retain patient-specific information in longitudinal brain MRI. Across five encoders and two T1-weighted datasets, we find that several representations support strong patient linkage and re-identification under the evaluated threat model, despite the absence of an identification objective during encoder training. We also show that released embeddings retain information predictive of the subject's recorded sex, extending the privacy risk beyond identity linkage alone. These findings indicate that releasing learned representations instead of raw brain MRI does not, by itself, eliminate linkage or attribute-inference risk and is therefore insufficient on its own to establish anonymity.

Our study is limited to two longitudinal T1-weighted datasets, five pretrained encoders, and one inferred attribute. Moreover, the evaluated models differ simultaneously in architecture, pretraining domain, and objective, so our comparisons should not be interpreted as isolating the causal effect of any single design choice. Future work should move from measuring privacy risk to reducing it through privacy-aware training objectives and post-hoc sanitization methods that suppress identity-bearing information in released embeddings. An important next step is to evaluate how effectively these approaches reduce privacy risks while preserving downstream utility.

\subsection*{AI use statement}

Generative AI tools were used in this work to support language refinement and improve the clarity of the manuscript, as well as to assist with selected parts of the experimental implementation and figure generation. These tools were not used to define the research questions, perform the literature review, or establish the scientific conclusions of the study. All AI-assisted text and code were manually reviewed by the authors, and the resulting experiments and visualizations were independently checked for correctness and consistency. The authors remain fully responsible for the content of the manuscript, the reported results, and all material developed with the assistance of generative AI.



\subsection*{Ethics statement}
This work studies privacy risks in released brain MRI representations using existing research datasets and does not attempt to identify participants in the real world. Our re-identification evaluation is performed at the dataset-subject level: the experiments measure whether different scans can be linked to the same subject, while attribution to a real-world identity is considered only as an explicit assumption in the threat model when identity-linked auxiliary information is available. We do not recover names or other civil identifiers, nor do we release patient-level linkage results. The purpose of the study is to characterize potential privacy risks and inform safer sharing of medical-image representations; all reported results are presented in aggregate.



\subsection*{Reproducibility statement}
We provide the code required to reproduce the feature-extraction, linkage, re-identification, attribute-inference, and evaluation pipelines in an anonymous repository: {\url{https://anonymous.4open.science/r/mri-reid-privacy-86C3/README.md}}. The repository includes the model-specific  embedding-extraction procedures, evaluation scripts, metric implementations, and plotting code used in the experiments. The datasets evaluated in this work are publicly available from their original providers and are not redistributed. Additional implementation details, including model configurations, data-processing choices, and ablation settings, are documented in the appendix and repository.



\bibliography{iclr2027_conference}
\bibliographystyle{iclr2027_conference}
\newpage
\appendix
\section{Additional Results} 

\subsection{Ablation: Distance Metric} \label{app:ablation_distance_metric}

To assess the sensitivity of our results to the similarity function, we compare Canberra-based and cosine similarity in Table~\ref{tab:canberra-vs-cosine}. Canberra yields higher linkability for every encoder on both datasets, indicating greater separation between the same-subject and different-subject score distributions under this metric. Rank-1 matching is less sensitive to this choice: neither similarity function consistently dominates across all model--dataset combinations, although Canberra achieves slightly higher performance overall. We therefore retain Canberra for the main analysis because it consistently provides the stronger linkability audit while supporting comparable rank-1 matching results.

\begin{table}[ht]
  \centering
  \resizebox{\linewidth}{!}{%
  \begin{tabular}{lcccccccc}
    \toprule
     & \multicolumn{4}{c}{OASIS-2} & \multicolumn{4}{c}{MIRIAD} \\
    \cmidrule(lr){2-5} \cmidrule(lr){6-9}
     & \multicolumn{2}{c}{Linkability} & \multicolumn{2}{c}{\reidrate} & \multicolumn{2}{c}{Linkability} & \multicolumn{2}{c}{\reidrate } \\
    \cmidrule(lr){2-3} \cmidrule(lr){4-5} \cmidrule(lr){6-7} \cmidrule(lr){8-9}
    Model & Canberra & Cosine & Canberra & Cosine & Canberra & Cosine & Canberra & Cosine \\
    \midrule
    AnatCL          & \textbf{0.9312} & 0.5707 & \textbf{0.9205} & 0.8716 & \textbf{0.9714} & 0.6017 & \textbf{0.9971} & \textbf{0.9971} \\
    3D-Neuro-SimCLR & \textbf{0.6571} & 0.3502 & \textbf{0.6647} & 0.4794 & \textbf{0.7938} & 0.3888 & \textbf{0.9787} & 0.9176 \\
    Brain-MRI-SSL   & \textbf{0.5647} & 0.4204 & 0.5764 & \textbf{0.6461} & \textbf{0.8460} & 0.6674 & \textbf{0.9816} & 0.9788 \\
    DINO-ViT        & \textbf{0.7796} & 0.5694 & 0.8235 & \textbf{0.8412} & \textbf{0.8761} & 0.8677 & 0.9870 & \textbf{0.9899} \\
    SAM-Med3D       & \textbf{0.2627} & 0.2177 & 0.1394 & \textbf{0.1528} & \textbf{0.5139} & 0.4691 & 0.8958 & \textbf{0.9085} \\
    \bottomrule
  \end{tabular}}
  \caption{Linkability and rank-1 matching under Canberra-based and cosine similarity. Bold marks the higher (i.e., stronger attack) value for each pair.}
  \label{tab:canberra-vs-cosine}
\end{table}

\subsection{Cross Dataset Attribute Inference}
\label{app:sex_cross}

In Section~\ref{sec:inference}, we showed that the evaluated embeddings retain information predictive of recorded sex when the classifier is trained and tested within the same dataset. We now examine whether this information generalizes across cohorts by training on one dataset and testing on the other. Table~\ref{tab:sex} reports results in both directions, where A$\rightarrow$B denotes training on dataset A and testing on dataset B. Balanced accuracy decreases for every encoder under cross-dataset evaluation, with most models approaching chance level ($0.5$) in at least one direction. SAM-Med3D is the main exception, retaining balanced accuracy of 0.81 for OASIS-2$\rightarrow$MIRIAD and 0.79 for MIRIAD$\rightarrow$OASIS-2, compared with 0.86--0.89 within datasets. Interestingly, this differs from the linkage results, where SAM-Med3D exhibited the lowest linkability and rank-1 matching among the evaluated encoders, further showing that different privacy risks need not induce the same ordering across representations. Cross-dataset performance is also asymmetric: training on MIRIAD and testing on OASIS-2 performs better than the reverse for all encoders except SAM-Med3D. This asymmetry suggests sensitivity to dataset-specific factors such as preprocessing, acquisition conditions, and cohort characteristics; however, our experiments do not isolate which of these factors drives the observed domain shift.

\begin{table*}[t]
  \centering
  \resizebox{\linewidth}{!}{%
  \begin{tabular}{l cc cc | cc cc}
    \toprule
    & \multicolumn{4}{c|}{\textbf{Within-dataset}} & \multicolumn{4}{c}{\textbf{Cross-dataset}} \\
    \cmidrule(lr){2-5} \cmidrule(lr){6-9}
    & \multicolumn{2}{c}{OASIS-2} & \multicolumn{2}{c|}{MIRIAD}
    & \multicolumn{2}{c}{OASIS-2 $\rightarrow$ MIRIAD} & \multicolumn{2}{c}{MIRIAD $\rightarrow$ OASIS-2} \\
    \cmidrule(lr){2-3} \cmidrule(lr){4-5} \cmidrule(lr){6-7} \cmidrule(lr){8-9}
    Model & Bal. Acc. & Prec. & Bal. Acc. & Prec. & Bal. Acc. & Prec. & Bal. Acc. & Prec. \\
    \midrule
    AnatCL          & 0.6772 & 0.6965 & 0.6946 & 0.6983 & 0.4784 & 0.4769 & 0.5130 & 0.5132 \\
    3D-Neuro-SimCLR & 0.7011 & 0.7075 & 0.7789 & 0.7806 & 0.5000 & 0.2763 & 0.6170 & 0.6198 \\
    Brain-MRI-SSL   & \textbf{0.9093} & \textbf{0.9153} & \textbf{0.9546} & \textbf{0.9568} & 0.5032 & 0.7783 & 0.6351 & 0.7759 \\
    DINO-ViT        & 0.7894 & 0.8100 & 0.8716 & 0.8704 & 0.5151 & 0.6039 & 0.5464 & 0.7682 \\
    SAM-Med3D       & 0.8632 & 0.8735 & 0.8850 & 0.8887 & \textbf{0.8098} & \textbf{0.8221} & \textbf{0.7859} & \textbf{0.8168} \\
    \bottomrule
  \end{tabular}}
  \caption{Sex inference from frozen embeddings. Within-dataset results use a patient-disjoint split; cross-dataset results train on one cohort and test on the other. Best value per column in bold.}
  \label{tab:sex}
\end{table*}

\subsection{Ablation: Impact of Number of Scans}
\label{app:nb-scans}
MIRIAD contains more repeated scans per subject than OASIS-2, which may increase the success of a retrieval-based matching attack by providing more genuine references in the gallery. To assess the effect of this difference, we repeat the MIRIAD analysis after randomly subsampling each subject to between 2 and 5 scans, matching the range observed in OASIS-2. We repeat the subsampling ten times and report the mean and standard deviation for each model.

\begin{table}[ht]
  \centering
\begin{tabular}{lcccc}
    \toprule
     & \multicolumn{2}{c}{MIRIAD (full)} & \multicolumn{2}{c}{MIRIAD (capped 2--5)} \\
    \cmidrule(lr){2-3} \cmidrule(lr){4-5}
    Model & Linkability & \reidrate & Linkability & \reidrate \\
    \midrule
    AnatCL          & 0.9714 & 0.9971 & 0.9653 $\pm$ 0.0143 & 0.9883 $\pm$ 0.0064 \\
    3D-Neuro-SimCLR & 0.7938 & 0.9787 & 0.7829 $\pm$ 0.0214 & 0.9160 $\pm$ 0.0138 \\
    Brain-MRI-SSL   & 0.8460 & 0.9816 & 0.8420 $\pm$ 0.0166 & 0.9343 $\pm$ 0.0157 \\
    DINO-ViT        & 0.8761 & 0.9870 & 0.8778 $\pm$ 0.0182 & 0.9675 $\pm$ 0.0085 \\
    SAM-Med3D       & 0.5139 & 0.8958 & 0.5204 $\pm$ 0.0185 & 0.6246 $\pm$ 0.0298 \\
    \bottomrule
\end{tabular}
  \caption{Re-identification risk on MIRIAD: full dataset vs.\ capped (2--5 scans per subject). Capped results are reported as mean $\pm$ std over 10 random subsamplings.}
  \label{tab:reid-results_capped}
\end{table}

Table~\ref{tab:reid-results_capped} shows that capping the number of scans lowers the re-identification rate for all models, but to very different extents. AnatCL, DINO-ViT, and Brain-MRI-SSL are only mildly affected, with drops of 0.9, 2.0, and 4.7 percentage points, respectively. 3D-Neuro-SimCLR drops by 6.3 points and SAM-Med3D, already the least risky model, drops by far the most, by 27.1\% (from 89.58\% to 62.46\%). Linkability, in contrast, remains almost unchanged for all models except 3D-Neuro-SimCLR, varying by less than 0.01. Although SAM-Med3D loses 27.1 points in re-identification rate, its linkability remains essentially stable. 
Unlike rank-1 matching, linkability measures the separation between empirical same-subject and different-subject score distributions rather than retrieval success from a gallery. Subsampling can nevertheless alter these distributions by changing which acquisitions and pairwise comparisons are retained.

Two factors may explain why MIRIAD remains riskier than OASIS-2 even after capping the number of scans. First, MIRIAD scans of the same subject are acquired close together in time and are therefore very similar, whereas OASIS-2 scans are spaced at least one year apart. Second, all MIRIAD scans are released without preprocessing. OASIS-2 is released already preprocessed (motion-corrected, registered, and skull-stripped). This prior preprocessing may remove identifying information that is still present in the raw MIRIAD scans.

The number of scans per subject therefore contributes to re-identification, but other dataset-specific factors also shape the results. This does not change our overall conclusion:  re-identification remains high for all models, confirming our finding that the embeddings of these pretrained encoders are not privacy-safe.

\subsection{Score Distributions and Linkability}
\label{app:linkability}

\begin{figure}[ht]
  \centering
  \begin{subfigure}[b]{0.45\textwidth}
    \includegraphics[width=\linewidth]{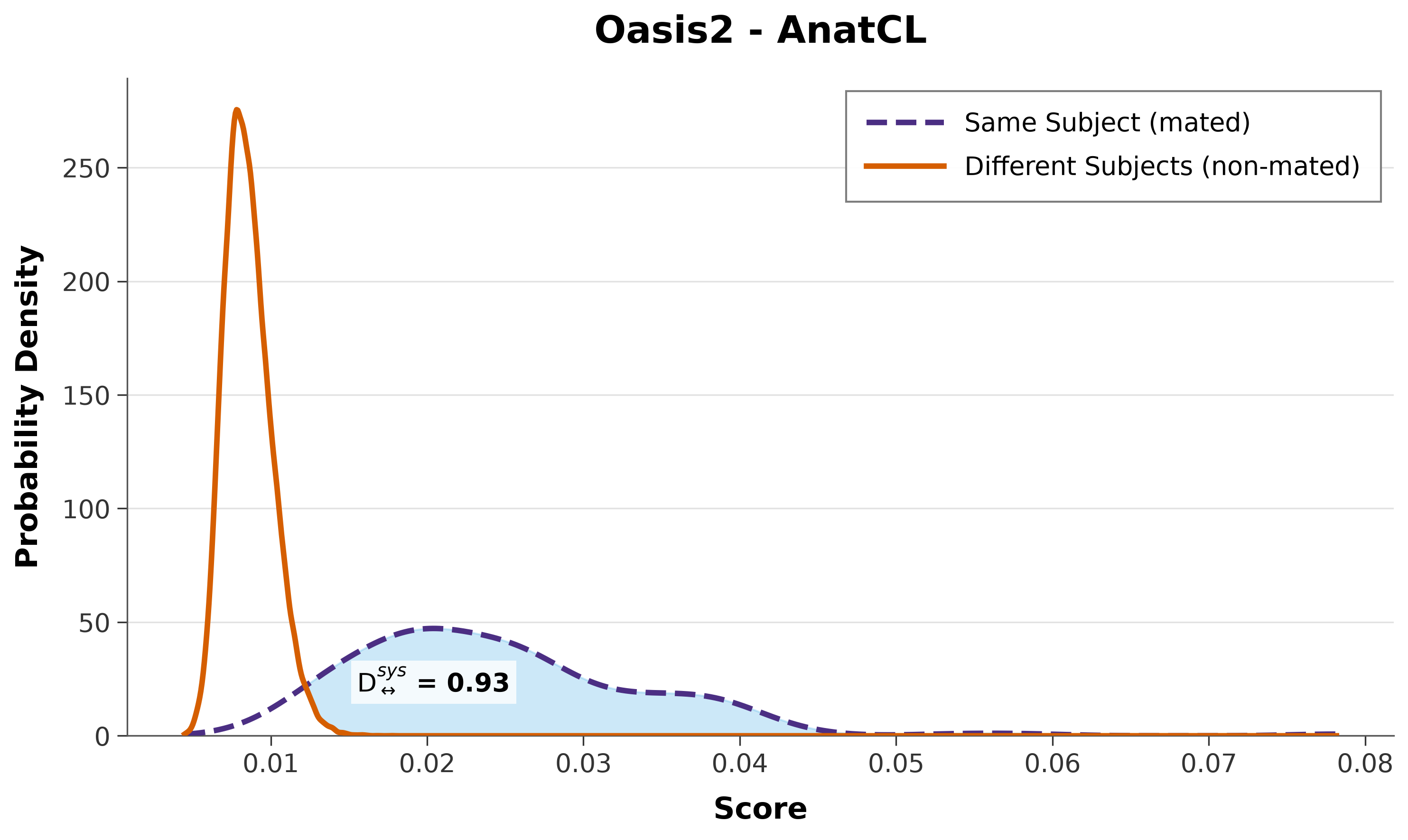}
    \caption{AnatCL, OASIS-2}
  \end{subfigure}\hfill
  \begin{subfigure}[b]{0.45\textwidth}
    \includegraphics[width=\linewidth]{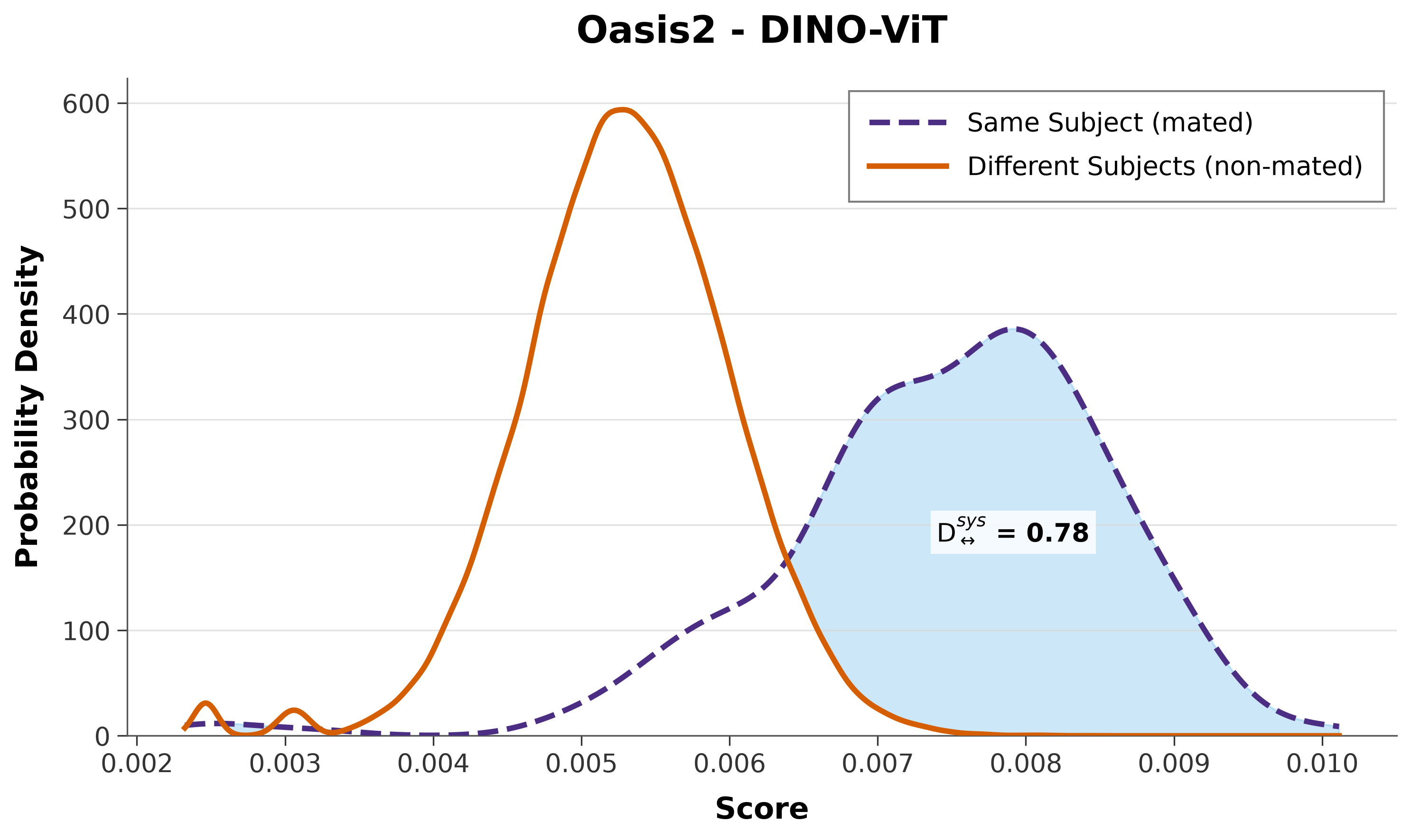}
    \caption{DINO-ViT, OASIS-2}
  \end{subfigure}

  \vspace{0.5em}
  \begin{subfigure}[b]{0.45\textwidth}
    \includegraphics[width=\linewidth]{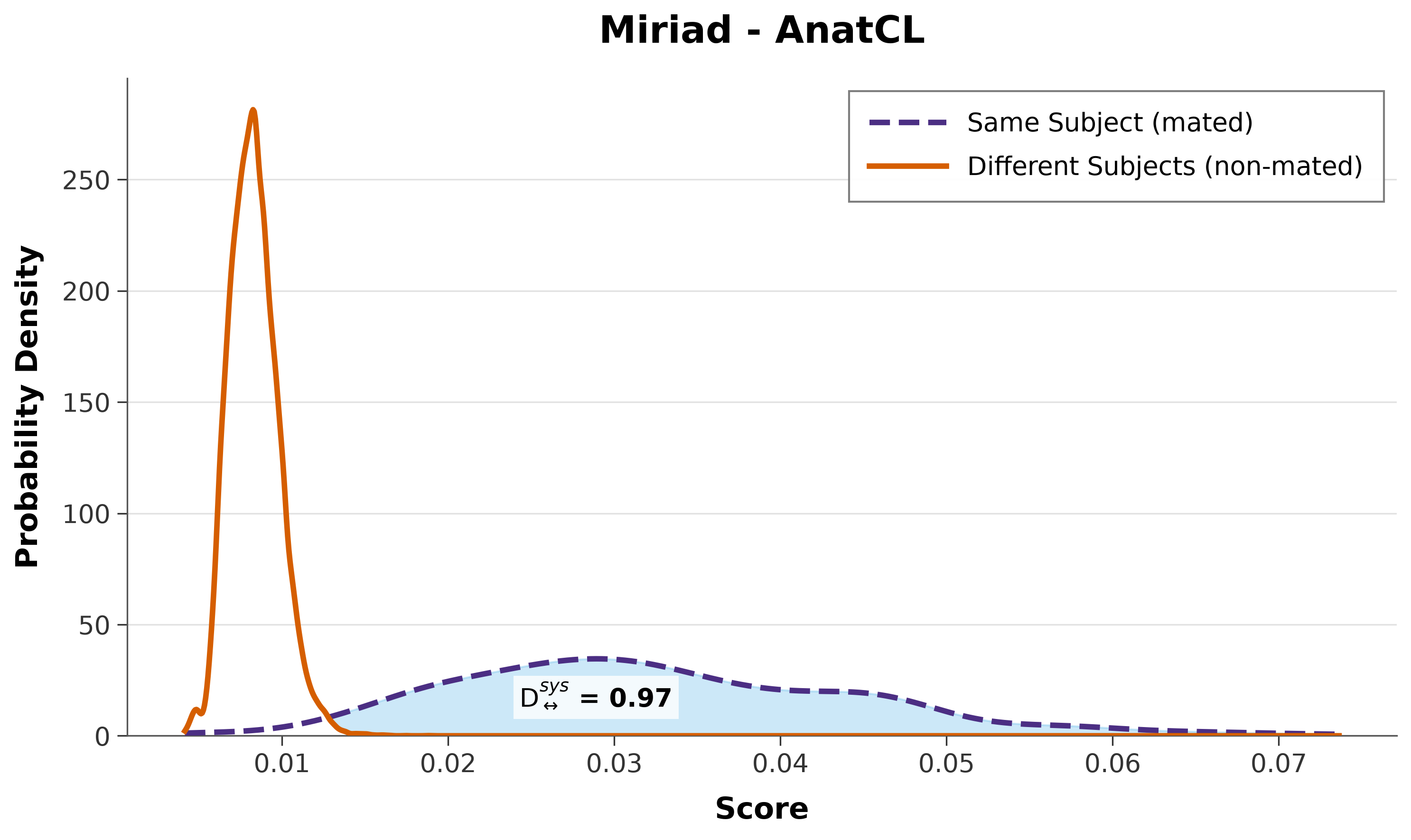}
    \caption{AnatCL, MIRIAD}
  \end{subfigure}\hfill
  \begin{subfigure}[b]{0.45\textwidth}
    \includegraphics[width=\linewidth]{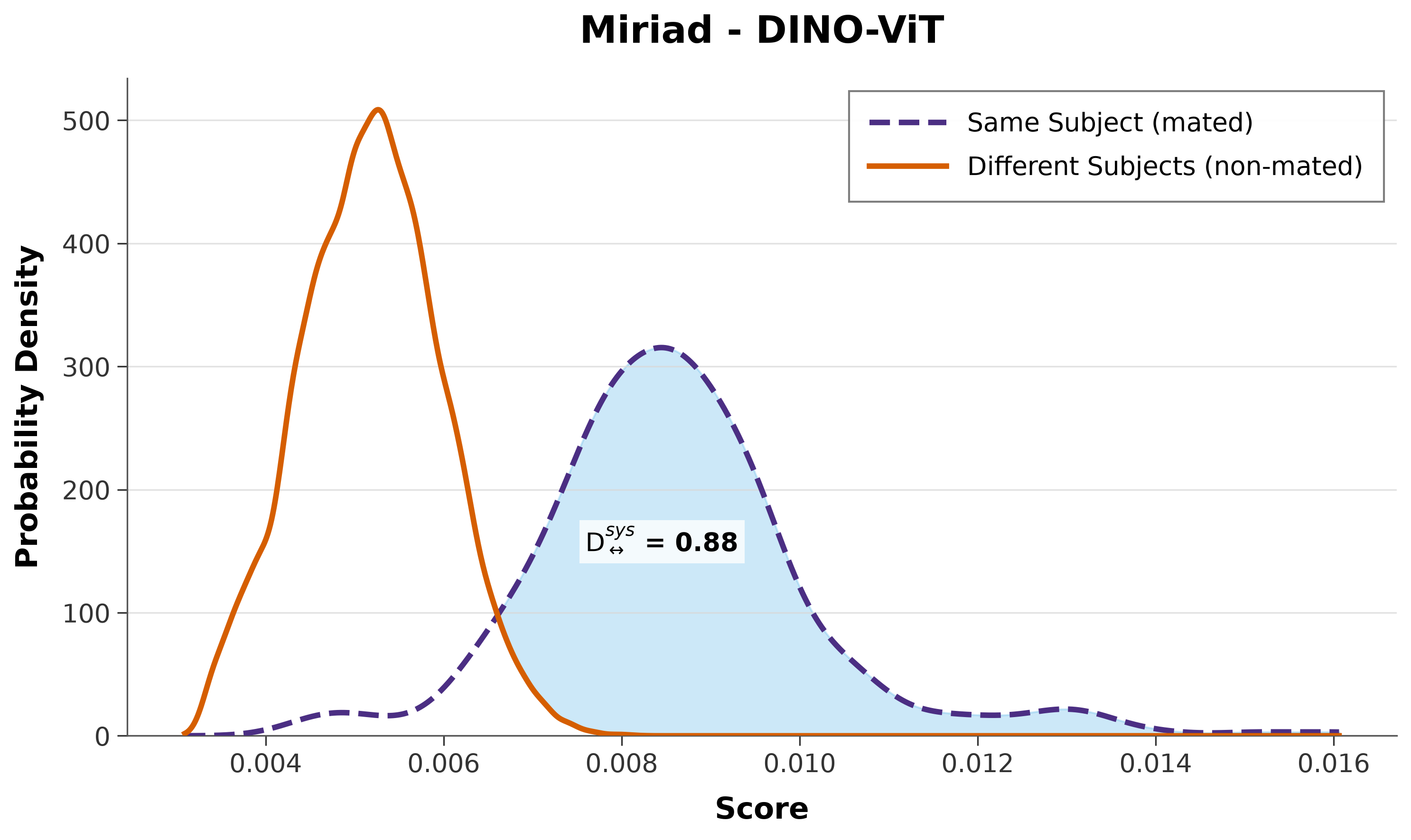}
    \caption{DINO-ViT, MIRIAD}
  \end{subfigure}
  \caption{Representative mated and non-mated score distributions (KDE) for AnatCL and DINO-ViT on OASIS-2 (top) and MIRIAD (bottom), with the linkability $D_{\leftrightarrow}^{\mathrm{sys}}$.}
  \label{fig:link_examples}
\end{figure}

Figure~\ref{fig:link_examples} shows representative distributions of Canberra similarity scores for mated (same-subject) and non-mated (different-subject) pairs for AnatCL and DINO-ViT on both datasets. The shaded region indicates where the mated density exceeds the non-mated density, and the corresponding system-level linkability $D_{\leftrightarrow}^{\mathrm{sys}}$ is reported in each panel.

\end{document}